\def\year{2021}\relax
\documentclass[letterpaper]{article} 
\usepackage{aaai21}  
\usepackage{times}  
\usepackage{helvet} 
\usepackage{courier}  
\usepackage[hyphens]{url}  
\usepackage{graphicx} 
\usepackage{natbib}  
\usepackage{caption} 
\usepackage{latexsym}
\usepackage{multirow}
\usepackage{amsmath}
\usepackage{amssymb}

\title{Integrating Pre-trained Model into Rule-based Dialogue Management}
\author {

        Jun Quan\textsuperscript{\rm 1,2}\thanks{Work done while Jun Quan was a research intern in Microsoft. Qiang Gan is the corresponding author (qigan@microsoft.com).},
        Meng Yang\textsuperscript{\rm 2},
        Qiang Gan\textsuperscript{\rm 2}, 
        Deyi Xiong\textsuperscript{\rm 1}, 
        Yiming Liu\textsuperscript{\rm 2}, Yuchen Dong\textsuperscript{\rm 2}, Fangxin Ouyang\textsuperscript{\rm 2}, Jun Tian\textsuperscript{\rm 2}, Ruiling Deng\textsuperscript{\rm 2}, Yongzhi Li\textsuperscript{\rm 2}, Yang Yang\textsuperscript{\rm 2} and Daxin Jiang\textsuperscript{\rm 2}\\
}

\affiliations {
    \textsuperscript{\rm 1} School of Computer Science and Technology, Soochow University, Suzhou, China \\
    \textsuperscript{\rm 2} STCA NLP Group, Microsoft\\
    terryqj0107@gmail.com, dyxiong@suda.edu.cn, $\{$menyan, qigan, yiml, yucdong, faouya, juti, ruilingd, yongli, yayan, djiang$\}$@microsoft.com
}
\begin{document}

\maketitle

\begin{abstract}
Rule-based dialogue management is still the most popular solution for industrial task-oriented dialogue systems for their interpretablility. However, it is hard for developers to maintain the dialogue logic when the scenarios get more and more complex. On the other hand, data-driven dialogue systems, usually with end-to-end structures, are popular in academic research and easier to deal with complex conversations, but such methods require plenty of training data and the behaviors are less interpretable. In this paper, we propose a method to leverages the strength of both rule-based and data-driven dialogue managers (DM). We firstly introduce the DM of Carina Dialog System (CDS, an advanced industrial dialogue system built by Microsoft). Then we propose the ``model-trigger'' design to make the DM trainable thus scalable to scenario changes. Furthermore, we integrate pre-trained models and empower the DM with few-shot capability. The experimental results demonstrate the effectiveness and strong few-shot capability of our method.
\end{abstract}

\section{Introduction}
Task-oriented dialogue systems usually consist of three main components, including natural language understanding, dialogue management, and natural language generation. The Dialogue Manager (DM) keeps tracking the current dialogue state and determines the next action to be taken. 

Rule-based DMs are widely used in industrial task-oriented dialogue systems, such as Google’s Dialog Flow and Microsoft’s Bot Framework. However, due to manual rules and rigid structure of dialogue flow, it's hard for developers to maintain a rule-based DM, especially when the scenario is complex. In recent years, data-driven approaches for task-oriented dialogue, usually with end-to-end architectures, have been an active area in the research community. \citet{wu2020tod} propose a task-oriented dialogue BERT (ToD-BERT), which is trained on 9 English task-oriented datasets across over 60 domains and achieves good performance on 4 dialogue subtasks.

We are motivated to leverage both advantages of the methods from industrial production and academic research. We hope to effectively reduce the cost in manually developing DM rules while ensuring the performance.

Based on CDS, we maintain its basic structure and incorporate the data-driven methods into the DM module. Specifically, we maintain the core architecture (i.e. Event-Trigger-Action mechanism) of the rule-based solution of the original DM to inherit its controllability, and take ``model-trigger'' design to replace original expression triggers. Furthermore, we integrate a task-oriented dialogue pre-trained model to make full use of the rich external linguistic knowledge learned from pre-training.

The contributions of our work are as follows. 1) We introduce an advanced dialogue management scheme of the industry-scale Carina Dialog System. 2) We introduce the ``model-trigger'' design in rule-based DM to achieve data-driven capability. 3) We propose an approach to integrating pre-trained model to DM. 4) We define dialogue actions in CDS based on CamRest676 dataset and re-annotate the dataset to evaluate our method. 5) We design experiments to prove the feasibility and strong few-shot capability of our method. 6) We build a demonstration system to visualize the process of our model-trigger DM \footnote{Link to our video: \url{https://youtu.be/suQTJ-L_3j4}}.

\begin{figure}[t] 
\centering 
\includegraphics[scale=0.3]{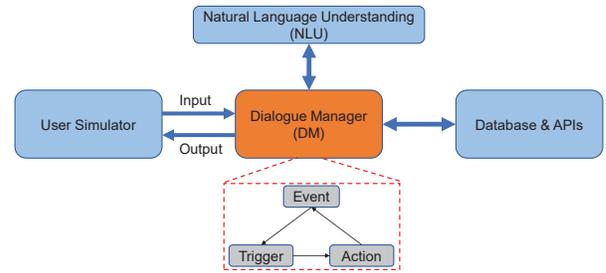}
\caption{The architecture of Carina Dialog System}
\vspace{-1em}
\label{carina dialog system figure} 
\end{figure}

\begin{figure*}[t] 
\centering 
\includegraphics[scale=0.52]{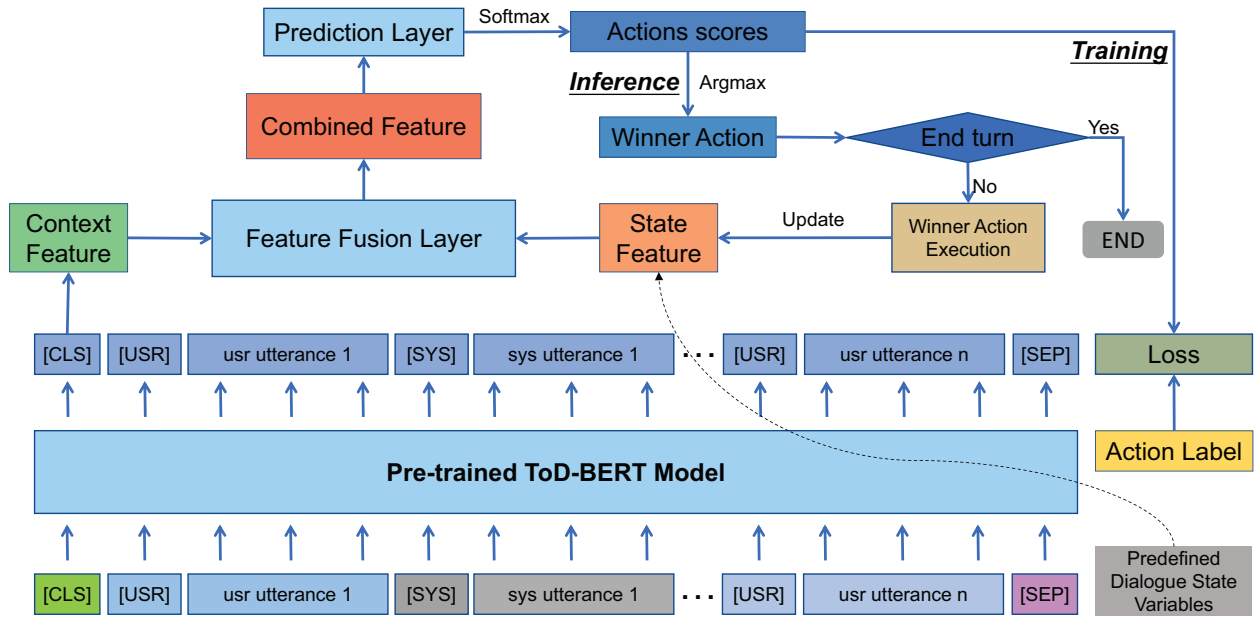}
\caption{The architecture of Model-Trigger DM}
\vspace{-1em}
\label{model trigger figure} 
\end{figure*}

\section{Method}
\textbf{Carina Dialog System (CDS)}. It is an industrial task-oriented dialogue system, which has been applied to many real-world scenarios. 
The overall architecture of CDS is shown in Figure \ref{carina dialog system figure}. It consists of four main components: 1) a User Simulator module; 2) a Natural Language Understanding (NLU) module; 3) a Dialogue Manager (DM); and 4) a Database \& APIs module. In our work, we focus on the DM which decides next actions according to dialogue state. 

\textbf{Event-Trigger-Action}. This is the core mechanism of the DM. The system receives an external event (e.g. $Start$, $Query$ or $End$) when each turn starts. All triggers will be evaluated when new events come. If the triggering condition is satisfied for a certain trigger, it will be activated and its corresponding action will execute. Then dialogue state will be updated accordingly and system will generate new internal events. The ``event-trigger-action'' process is running as a loop until no trigger is activated any more. A winner action sequence (containing one or more actions) is selected out when a turn ends. We refer to the process of predicting each action as mini-turn. 

The original trigger module of CDS is implemented through the expression trigger, where firing conditions are formed with manually written expressions composed of dialogue state variables, which can ``listen to'' various customized events. Developers can build flexible and interpretable dialogue logics with expressions. However, it is hard for them to maintain these logic rules when the scenario becomes complex. So we propose the ``model-trigger'' design to achieve data-driven capability and reduce developers' efforts to write expressions.

\textbf{Model-Trigger DM}. We are motivated to leverage a neural model to predict the winner action of each mini-turn according to not only the dialogue state used in expression triggers, but also the dialogue context. Thus, we hope the model trigger can cover most of the manually defined expressions. The architecture of our model-trigger DM is shown in Figure \ref{model trigger figure}. In each mini-turn, the model takes two parts of features to predict a winner action, including context feature and state feature. 1) \textbf{Context Feature:} All utterances in dialogue context are concatenated into a flat sequence as the input. We use ToD-BERT to encode the context sequence and take the embedding of the ``[CLS]'' token as the context feature. 2) \textbf{State Feature:} We treat the binarization of pre-defined dialogue state variables during the process, which updates in each mini-turn, as the state feature.

We formulate the action prediction as a multi-class classification problem. The two parts of features are combined through a feature fusion layer and used to predict the winner action of each mini-turn over all possible actions through a prediction layer. Both the feature fusion layer and the prediction layer are implemented with deep neural networks.

To evaluate our method, we create a new version of CamRest676 dataset \cite{wen2016conditional,wen2016network} and conduct experiments on it. We define executable dialogue actions and annotate each turn of all dialogues with dialogue action labels. Each turn may contain a sequence of action labels, which means the system response in this turn is generated during the execution of those actions in order. Then we feed the dialogue sessions with labels into CDS and run in a ``data-collection mode''. In this mode, DM will execute the labeled actions in order to update the internal dialogue state features, while dialogue context features are fixed for each turn. Each turn may produce more than one training data records, corresponding to the number of action labels. Experiments show that it can achieve 95.73\% of mini-turn accuracy to predict dialogue action, which is a great performance. Furthermore, few-shot experiments show a satisfactory performance using only small-scale training data.

\begin{figure*}[t] 
\centering 
\includegraphics[scale=0.63]{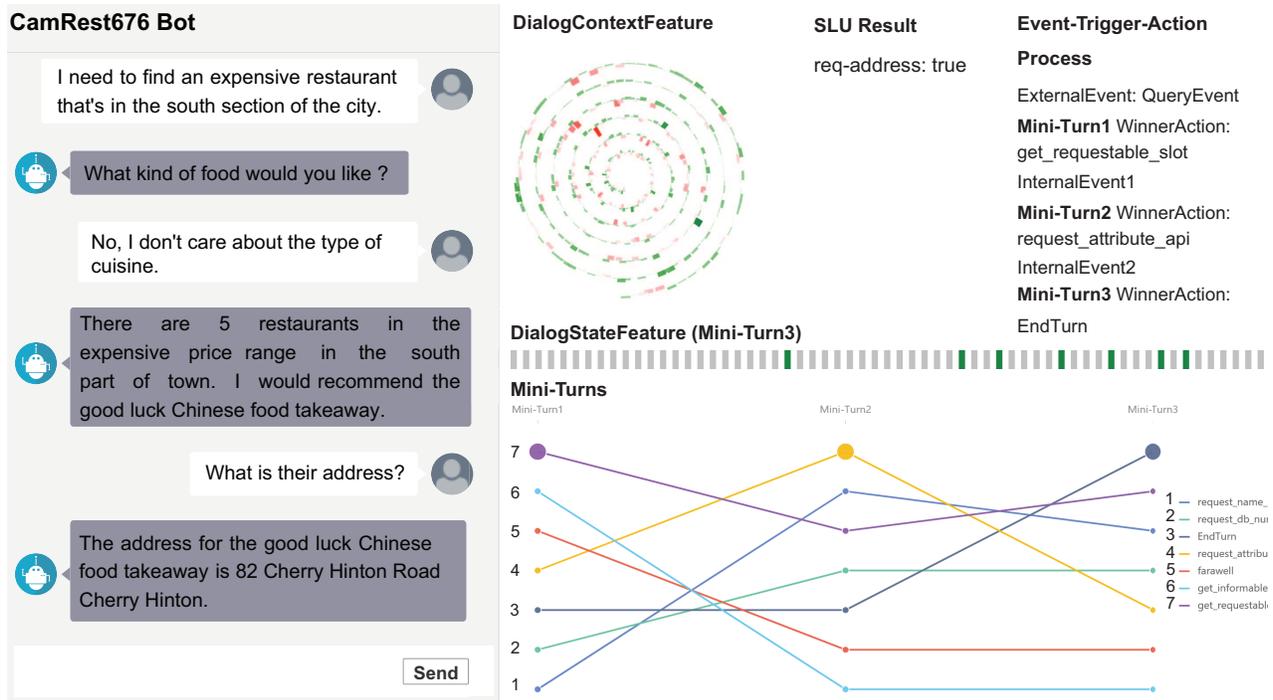}
\caption{Demo showing the process of Model-Trigger DM.}
\vspace{-1em}
\label{demo} 
\end{figure*}

\section{Demonstration}
As shown in Figure \ref{demo}, we build a demonstration system based on our model-trigger DM. The main function of our system is to show the process of predicting dialogue actions according to the dialogue context and the current state. On the left side of the interface are the dialogue context display area and input area. On the right side, the results of natural language understanding in each turn are shown in the form of semantic frame at the top part. We visualize the 768-dimensional dialogue context feature and the 64-dimensional dialogue state feature in each mini-turn. The probability distribution of predicting dialogue actions in each mini-turn is shown at the bottom part. Through the demonstration system, we can clearly and intuitively observe the whole process of predicting dialogue actions by our model-trigger DM.

\section{Conclusion and Future Work}
In this paper, we introduce Carina Dialog System and propose ``model-trigger'' to reduce human efforts. We further integrate a pre-trained model for few-shot capability. Experiments show the good performance of our method. We build a demonstration system to visualize the process of the model-trigger DM. In our future work, we will further improve the model-trigger DM and apply it to the real-world products to reduce the manual efforts of traditional solution.

\section{Acknowledgments}
This work is supported by Microsoft STCA NLP Group. We would also like to thank the anonymous reviewers for their insightful comments.

\bibliography{aaai2021.bib}

\end{document}